\documentclass[10pt, journal]{IEEEtran}
\usepackage{amsmath,amsfonts}
\usepackage{wasysym}
\usepackage{amsmath,amsfonts}
\usepackage{amsmath}
\usepackage{amssymb}
\usepackage{xfrac}
\usepackage{soul}
\usepackage{multirow} 
\usepackage{tikz}
\usepackage{pifont} 

\usepackage[ruled,linesnumbered]{algorithm2e}
\usepackage{algpseudocode}
\usepackage{textcomp}
\usepackage{stfloats}
\usepackage{verbatim}
\usepackage{graphicx}
\usepackage[numbers,sort&compress]{natbib}
\usepackage{array}
\usepackage{bm}
\usepackage{balance}

\usepackage{color,xcolor}
\usepackage{booktabs}
\usepackage{array}
\usepackage{float}
\usepackage{makecell}
\usepackage{tabularx}
\usepackage{pifont}
\usepackage{circledsteps}

\makeatletter
\usepackage{tikz}

\newcommand{\figref}[1]{Fig.\;\ref{#1}}
\newcommand{\tabref}[1]{Table \ref{#1}}

\soulregister{\cite}7 
\soulregister{\citep}7 
\soulregister{\citet}7 
\soulregister{\ref}7 
\soulregister{\pageref}7 
\soulregister{\figref}7
\soulregister{\tabref}7

\newcommand\uhl[2][\usrcolor]{\sethlcolor{#1}\hl{#2}}

\begin{document}





\title{\uhl{StereoTacTip: Vision-based Tactile Sensing with Biomimetic Skin-Marker Arrangements}}

\author{
        \IEEEauthorblockN{Chenghua~Lu$^{1}$, Kailuan~Tang$^{2}$, Xueming Hui$^{3}$, Haoran Li$^{4}$, Saekwang Nam$^{5}$, Nathan F. Lepora$^{1}$}
        \thanks{This work was supported by a UKRI award ``Research Centre for Smart, Collaborative Industrial Robotics" under Grant EP/V0612158/1, and a China Scholarship Council-University of Bristol Joint Scholarship under Grant 202209370035. (Corresponding author: Nathan F. Lepora. Email: {\rm\footnotesize n.lepora@bristol.ac.uk}.)}
        \thanks{\IEEEauthorblockA{$^{1}$Chenghua Lu and Nathan F. Lepora are with the School of Engineering Mathematics and Technology and Bristol Robotics Laboratory, University of Bristol, Bristol, U.K.}
        $^{2}$Kailuan Tang is with the School of Mechatronics Engineering, Harbin Institute of Technology, Harbin, China. 
        {$^{3}$Xueming Hui is with the Fuyao Institute for Advanced Study, Fuyao University of Science and Technology, Fuzhou, China. }
        {$^{4}$Haoran Li is with School of Robotics, Xi'an Jiaotong-Liverpool University, Suzhou, China.}
        {$^{5}$Saekwang Nam is with the Graduate School of Data Science at Kyungpook National University, Daegu, Republic of Korea (South). }}
        }
\maketitle


\begin{abstract} 
Vision-Based Tactile Sensors (VBTSs) stand out for their superior performance due to their high-information content output. Recently, marker-based VBTSs have been shown to give accurate geometry reconstruction when using stereo cameras. \uhl{However, many marker-based VBTSs use complex biomimetic skin-marker arrangements, which presents issues for the geometric reconstruction of the skin surface from the markers}. Here we investigate how the marker-based skin morphology affects stereo vision-based tactile sensing, using a novel VBTS called the StereoTacTip. To achieve accurate geometry reconstruction, we introduce: (i) stereo marker matching and tracking using a novel Delaunay-Triangulation-Ring-Coding algorithm; (ii) a refractive depth correction model that corrects the depth distortion caused by refraction in the internal media; (iii) a skin surface correction model from the marker positions, relying on an inverse calculation of normals to the skin surface; and (iv)~methods for geometry reconstruction over multiple contacts. To demonstrate these findings, we reconstruct topographic terrains on a large 3D map. \hl{Even though contributions (i) and (ii) were developed for biomimetic markers, they should improve the performance of all marker-based VBTSs.} Overall, this work illustrates that a thorough understanding and evaluation of the morphologically-complex skin and marker-based tactile sensor principles are crucial for obtaining accurate geometric information.
\end{abstract}



\section{Introduction}

\IEEEPARstart{B}y transducing physical contact into informative data, tactile sensors enable robots to gather information about the geometric characteristics of their interacting environment~\cite{LuoReview, hardware}. Vision-Based Tactile Sensors (VBTSs) stand out for their superior performance due to their high-resolution output, enabling high-information content data acquisition over many pixels in a tactile image~\cite{CameraBased}. In particular, photometric VBTSs such as the GelSight~\cite{gelsight1} are inherently suited to reconstructing spatial geometry due to their use of an internal reflective layer that is sensitive to changes in depth ~\cite{markeror,GelsightReview}. In contrast, marker-based VBTSs do not possess this innate depth-sensing ability, but have other benefits such as being highly sensitive to shear~\cite{John}. However, in recent years, researchers have demonstrated that marker-based VBTSs can also achieve accurate geometry reconstruction with the use of multiple internal cameras~\cite{ShixinReview,TacLink, GelStereoBioTip,BiTac}, which is an ongoing topic of investigation that will benefit progressively more from the continuing miniaturization of camera technology.

\begin{figure}[!t]
 \vspace{0pt}
 \centering
 \includegraphics[width=3in]{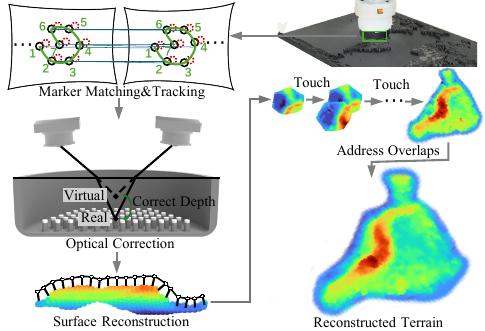}
 \caption{An overview of StereoTacTip's reconstruction process when touching the topographic model of South America.}
 \label{fig_overall}
 \vspace{-0.5em}
\end{figure}


\uhl{Meanwhile, an emerging trend for marker-based VBTSs is to improve the interpretability of contact features through the design of complex biomimetic skins, incorporating internal structures or contact-dependent marker properties~\cite{Bio-opticalReview}. The TacTip is a prototypical example of a marker-based VBTS with a biomimetic skin, by using marker-tipped pin-like structures on the underside of the skin, which act as mechanical amplifiers of skin deformation into larger marker displacements~\cite{TacTipFamily}. Versions with more complex biomimetic skin such as fingerprints to enhance incipient slip have also been proposed~\cite{FingerprintTacTip1,FingerprintTacTip2}. Sophisticated skins are also being adopted for other VBTSs, e.g. the ChromaTouch uses a double-layer coloured markers~\cite{ChromaTouch} and the BioTacTip uses markers that are uncovered with pressure~\cite{BioTacTip}.} Overall, the use of more `morphological intelligence' in VBTS skins is a promising direction where there is a scope for substantive innovation.

However, the use of greater morphological complexity in VBTS skins will cause issues for accurate geometry reconstruction using multiple internal cameras, which requires careful consideration to address. The issue is that depth perception is crucial for ensuring geometric accuracy, which in marker-based VBTSs can be obtained using binocular cameras to directly capture pairs of tactile images combined with stereo vision~\cite{TacLink, GelStereo}. But if the skin is morphologically complex, then there will be a complex relationship between the 3D position of the markers and the true deformation of the skin. Moreover, this relationship can become significantly more complicated when factors such as the refraction of light through the internal media of the VBTS are considered~\cite{GelStereoPalm,GelStereoBioTip}.

\begin{table*}[!t]
  \vspace{-19pt}
  {
  \centering
  \caption{Comparison of the proposed StereoTacTip with the existing state-of-the-art marker-based VBTS}
  \vspace{-3pt}
  \label{Tab1}
  \renewcommand{\arraystretch}{1.05}
   \begin{tabular}{@{}p{1.5cm} >{\centering\arraybackslash}p{1.67cm}>{\centering\arraybackslash}p{1.67cm}>{\centering\arraybackslash}p{2.2cm}>{\centering\arraybackslash}p{1.9cm}>{\centering\arraybackslash}p{1.5cm}>{\centering\arraybackslash}p{2.34cm}>{\centering\arraybackslash}p{2.2cm}c@{}}
  \toprule
  Sensor & TacTip~\cite{VGNN} & BioTacTip~\cite{BioTacTip} & TacLeg~\cite{TacLeg} & TacLink~\cite{TacLink}  & GelStereo~\cite{GelStereo} & GelStereo Palm~\cite{GelStereoPalm} & Our Work \\
  \hline
  Shape & dome & flat& leg-shape  & cylindrical & flat & dome & flat/customizable \\
  \hline
  Camera & monocular & monocular & monocular  & binocular& binocular & binocular & binocular \\
  \hline
  
  Marker & biomimetic tip &tip/cover  & simple dot  &simple dot  &simple dot  &simple dot   & biomimetic tip\\
  \hline

 Dimensionality & 2D &2.5D  &2.5D &3D &3D &3D &3D  \\
 \hline
  
  \begin{tabular}[l]{@{}l@{}}
  Depth Model\\(Accuracy)
  \end{tabular}
  &
  \begin{tabular}[c]{@{}c@{}}
  Voronoi diagram\\(+)
  \end{tabular}
  &
  \begin{tabular}[c]{@{}c@{}}
  linear intensity\\(+)
  \end{tabular}
  &
  \begin{tabular}[c]{@{}c@{}}
  perspective projection\\ (+)
  \end{tabular}
  &
  \begin{tabular}[c]{@{}c@{}}
  stereo triangulation\\ (++)
  \end{tabular}
  &
  \begin{tabular}[c]{@{}c@{}}
  stereo triangulation\\(++)
  \end{tabular}
  &
  \begin{tabular}[c]{@{}c@{}}
  ray tracking\\(+++)
  \end{tabular}
  &
  \begin{tabular}[c]{@{}c@{}}
  optical correction \\(+++)
  \end{tabular} \\
  \hline
  \begin{tabular}[l]{@{}l@{}}
  Matching/Tracking\\(Robustness)
  \end{tabular}
  & -- & -- & -- & 
  \begin{tabular}[c]{@{}c@{}}
  path tracking \\ (++)
  \end{tabular} 
   & 
  \begin{tabular}[c]{@{}c@{}}
  sorting/KNN  \\ (+)
  \end{tabular} 
  & 
  \begin{tabular}[c]{@{}c@{}}
  sorting/KNN \\(+)
  \end{tabular} 
  &
  \begin{tabular}[c]{@{}c@{}}
  DTRC \\(+++)  
  \end{tabular} \\
  \hline
  Simplicity & medium & medium & high & high & high & low & high\\
  \bottomrule
  \end{tabular}
  }
  \vspace{-15pt}
\end{table*}

\uhl{In this paper, we carefully address these issues of how to use a marker-based VBTS with a biomimetic complex skin to obtain accurate 3D geometry reconstruction. Based on the development of a novel stereo-tactile sensor, the StereoTacTip, equipped with a replaceable marker-based skin, we systematically highlight and address multiple issues for both the hardware and algorithmic approaches.} We ensure that the tactile sensors can provide accurate, real-time depth and geometry information, which can then be combined into a large surface reconstruction (\figref{fig_overall}). 

The main contributions of this article are as follows:

\noindent 1) \textbf{Stereo Marker Matching and Tracking Algorithm}:
We develop a novel Delaunay-Triangulation-Ring-Coding algorithm for the simultaneous implementation of marker matching and tracking across stereo tactile images. Specifically, our method utilizes a Delaunay-based approach to form a mesh of markers, then iterates through the ring-shaped edge markers to assign each a unique identifier. Overall, this algorithm enables highly-robust marker matching, and we demonstrate its effectiveness on several marker layouts. 
  
\noindent 2) \textbf{Refractive Depth Correction Model}:
The effects of refraction from the internal media of the VBTS are carefully investigated with a `skin-less' sensor, from which we discover and model an approximately linear relationship between the actual and virtual depth changes of a  marker and those observed by the camera. We use this model to correct the depth distortion caused by light refraction, including a calibration step that measures the refractive index of the internal media. 
  
\noindent 3) \textbf{Skin Surface Correction from Marker Positions}:
\uhl{The effects of the biomimetic complex skin -- here the pins -- on relating the marker displacements to the skin deformation is analytically modelled.} In particularly, we develop a model that inversely calculates the normals to the surface formed by the markers, then fitting a new skin surface displaced by the length of the pins along these normals. We develop tests to assess the accuracy of the 3D geometry reconstruction and its relation to skin thickness and marker density. 

\noindent 4) \textbf{Geometry Reconstruction over Multiple Contacts}:
We also develop a geometric method to reconstruct a large surface from multiple overlapping contacts with a stereo VBTS, including contiguity and mollifying operations. To demonstrate the relevance and applicability of our findings, we employ these to reconstruct topographic terrains on a 3D world map.

\uhl{We systematically evaluate the performance of the proposed methods using the StereoTacTip as a representative case study. Importantly, many aspects of the methodology, such as contributions (1,2), apply broadly to all marker-based VBTSs.}


\vspace{-2pt}
\section{Related Works}
\label{Related_Works}

Table I shows a comparison of studies of marker-based VBTSs that obtain depth information from the markers on the inner sensing surface. For monocular VBTSs, the key to achieving geometry reconstruction is to build mappings between 2D displacements of markers and 3D contact properties such as depth~\cite{2d-3d}. For example, area-based tactile features provided by the Voronoi diagram have been used to approximate contact depth within each tessellation region~\cite{VGNN}. Alternatively, methods that make use of changing marker shape are known as 2.5D~\cite{2d-3d}, such as the BioTacTip that integrates a sharp white tip surrounded by black cover tips to emphasize the contact shape~\cite{BioTacTip}. Another 2.5D method is to use a camera axis almost perpendicular to the skin surface, as in the TacLeg, to utilize the perspective projection to inform about depth~\cite{TacLeg}. However, as these methods rely on indirect features that correlate with depth, their ability to provide precise depth maps remains limited.
For binocular sensors, the 3D reconstruction can be achieved through stereo vision to reconstruct precise 3D depth and contact information. Most such VBTSs use a pair of cameras to replace the monocular camera ({\textit{e.g.}}, \cite{TacLink, GelStereo}) and calculate the depth information by stereo-vision triangulation. For example, using the GelStereo palm, light ray tracking calculations applied to birefringent layers give precise depths~\cite{GelStereoPalm}. However, while this can have good geometric reconstruction accuracy, it can be computationally complex to compute.

Marker matching is important to ensure a one-to-one correspondence between features in left and right images, allowing for consistent tracking of corresponding markers across frames by monitoring their positions. Existing marker matching methodologies are specifically designed for particular marker configurations, including foresight sorting for rectangular patterns~\cite{GelStereo}, and polar sorting for circular patterns~\cite{GelStereoPalm}. However, these approaches exhibit limited generalizability, rendering them ineffective upon alteration of the marker configurations. \uhl{Meanwhile, marker tracking generally relies on the k-nearest neighbor (KNN) method~\cite{GelStereo, GelStereoPalm}, but this method performs poorly in scenarios involving large or fast changes in tactile skin deformation, {\em e.g.} under large contact velocity.}

By contrast, the novel methodologies proposed in this paper endow our StereoTacTip with strong robustness and accuracy while having a computationally simple algorithm. We show the methods maintain effectiveness even with variations in sensor shape, marker pattern and marker quantities, indicating their future generalizability to other marker-based VBTS research.

\begin{figure*}[t]
 \vspace{-20pt}
 \centering
 \includegraphics[width=6.4in]{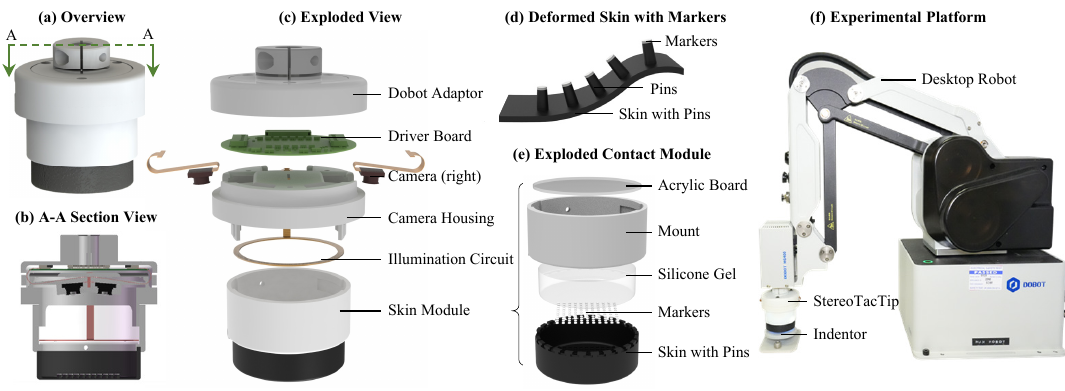}
\vspace{-5pt}
 \caption{From left to right: (a) - (c) illustrate the overall view, A-A section view and exploded view of StereoTacTip, respectively, while (e) further shows the exploded view of the skin module in (c). (d) shows the situation when the skin with pins (also presented in (e)) deforms. In the experimental platform demonstrated in (f), the StereoTacTip, mounted on the end of Dobot MG400, is pressing the indentor fixed to the desk.}
 \label{fig_design}
 \vspace{-1.2em}
\end{figure*}

\section{Methodology}
\label{Methodology}

\subsection{The StereoTacTip Design}
\label{StereoTacTip_Design}

The proposed StereoTacTip is a modular sensor that utilizes binocular vision to achieve stereo-tactile sensing (illustrated in~\figref{fig_design}). It consists of mechanical structural components for sensor mounting, housing, physical transduction, and embedded electronics for image acquisition and processing. 

\subsubsection{Mechanical Components}

The primary mechanical components include a connector for the robot arm, a camera housing, and a skin module, as shown in~\figref{fig_design}(c). The connector is specifically designed to interface with a desktop robot arm (MG400, Dobot Robotics Inc., CN), 
as shown in  \figref{fig_design}(f), 
the platform used in our experiments. 
The camera housing secures the driver board and the two cameras for stereo imaging, to ensure they are properly positioned and be stably affixed to the robot arm connector. As illustrated in~\figref{fig_design}(e), the skin module is fabricated using multi-material printing technology (J826 3D printer, Stratasys Inc., USA), \uhl{with the mount made of a rigid material (Polyjet resin VeroWhite, hardness: Shore D = 86), while the skin with pins made of a rubber-like soft material (Polyjet resin Agilus30, hardness: Shore A = 30). The markers (\diameter 1\,mm$\times$0.3\,mm), made in a contrasting color (white) and placed on the tips of pins (\diameter1.2\,mm$\times$1.5\,mm, 1$^\circ$$\times$1.5\,mm  chamfer on top), are used to predict the deformation of the skin (\figref{fig_design}(d)). A 1\,mm-thick acrylic board is glued to the inner pipe of the skin module, and the 10\,mm cavity inside the pipe is injected with optically clear silicone gel (RTV27905, Techsil Inc., UK, penetration: $3\sim7$ mm). Unless otherwise specified, the default markers' separation is 2.54\,mm, with a standard skin thickness of 1\,mm.}

\subsubsection{Electronic Components}

\uhl{The electronic components consist of two cameras (OV5693: 640$\times$480 pixels; $120^\circ$ field of view; 30 fps)}, a stereo processing circuit and an illumination circuit. The main control chip (SPCA2089B) of the driver board synchronously drives the cameras to capture images simultaneously and combines them into a single image frame. The cameras connect to the driver board via a 20-pin flexible printed circuit and are securely attached to the camera housing using double-side tape. The front of the driver board features a 4-pin socket for USB communication. The cameras capture left and right images, outputting them in a concatenated format of 1280$\times$480 pixels. The illumination circuit consists of a flexible ring with four white LEDs powered through a 10-pin connector located on the back of the driver board.


The experimental platform used for the stereo-tactile reconstruction tasks comprises the StereoTacTip mounted as the end-effector to the desktop robot arm and a test sample fixed to the manipulation platform  (\figref{fig_design}(f)). The sensor and arm are connected via USB to a PC (AMD Ryzen R7-5800X@4.5GHz, 32GB RAM). 
The reconstruction process will be detailed in Sections~\ref{Validations},~\ref{Evaluations} and~\ref{Demonstrations}.

\vspace{-10pt}
\subsection{Marker Localization}
\label{Marker_Localization}

\begin{figure}[t]
 \vspace{-2pt}
 \centering
 \includegraphics[width=3.1in]{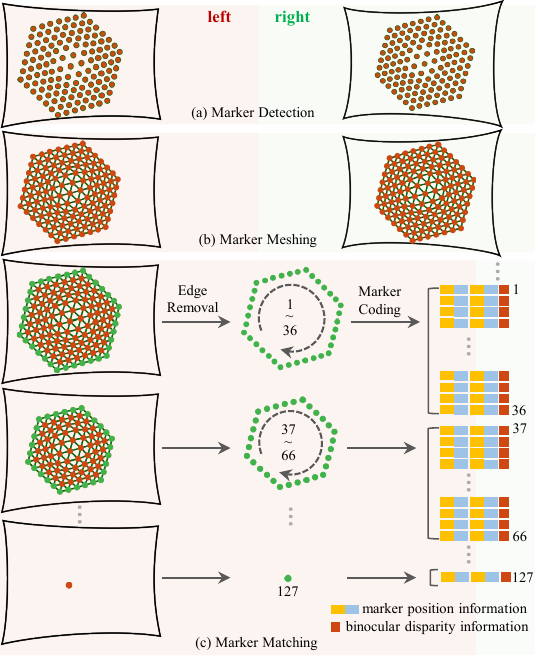}
 \caption{Process of marker matching (Steps 4-11 in Alg. 1). 
 This process is one part of 3D marker localization, following the calibration. 
 For simplicity, we only show the comprehensive process of a left image, and the initial and result of a right image. The output of DTRC is the markers' sequence, with marker position and binocular disparity information included.}
 \label{fig_matching}
 \vspace{0em}
\end{figure}

The 3D reconstruction of the skin deformed by contact relies on the precise 3D tracking of the marker array within StereoTacTip. We estimate the movement of each marker based on the principle of binocular disparity between the two cameras. The tracking and 3D reconstruction of the markers reduce computational complexity compared to the pixel-wise calculations, enabling faster updates of the tactile signals. The following subsections detail the data processing of marker positions for 3D skin deformation reconstruction.


\subsubsection{Camera Calibration}

Firstly, we calibrate the images captured separately by the binocular cameras using both intrinsic and extrinsic parameters. The calibration is performed using a high-precision chessboard made of an alumina plate (ZW Precision Machinery Co., CN) arranged in a 9$\times$7 layout, with each square measuring $4$\,mm per side. The 20 pairs of captured chessboard images are imported into a stereo camera calibrator (in MATLAB), resulting in a re-projection error of 0.12 pixels. Note that the skin module were not assembled during the calibration process. Consequently, this process did not account for the potential impact of the transparent elastomer layers' refraction on the calibration parameters (calibrated parameters summarized in Table~\ref{tab:camera_parameters}).

\begin{table}[!t]
\vspace{-5pt}
\centering
\caption{Binocular Camera Calibration Results}
\label{tab:camera_parameters}
\begin{tabular}{lcc}
\hline
\textbf{Camera} & \textbf{Left} & \textbf{Right} \\ 
\hline
Focal Length [pixel] & (442.37, 442.51) & (435.66, 435.76) \\ 
Principal Point [pixel] & (315.67, 237.97) & (324.72, 237.97) \\ 
Radial Distortion & (-0.74, -0.03) & (-0.07, -0.05) \\ 
Tangential Distortion & (0.00017, -0.00075) & (0.00060, -0.00064) \\ 
Rotation & \multicolumn{2}{c}{
$\begin{bmatrix}
0.9878 & 0.0106 & -0.1554 \\
-0.0104 & 0.9999 & 0.0022 \\
0.1554 & -0.0005 & 0.9879
\end{bmatrix}$} \\
Translation [mm] & \multicolumn{2}{c}{(12.49, -0.08, 0.59)} \\ 
\hline
\end{tabular}
\vspace{5pt}
\end{table}

\subsubsection{Marker Matching and Tracking} 


To detect the marker array in a single image (a resolution of 1280$\times$640 pixels), we first converted it to grayscale, then applied the Determinant of Hessian method for blob detection to identify the central positions of all markers in pixel coordinates~\cite{DexiTac}. Following this, we introduce a novel algorithm called Delaunay-Triangulation-Ring-Coding (DTRC) to encode and match the markers between the left and right images (Alg. 1). The algorithmic process of DTRC is described below:
\begin{algorithm}[t]
  \footnotesize
  \SetCommentSty{TimesNewRoman}

  \caption{Marker Detection and Matching}
  \SetAlgoRefName{Algorithm 1}
  \label{alg_grasping}

  \KwIn{$I_{n\_{\rm raw}}, n=1,2$ \tcp*[r]{captured raw images from cameras}}
  \KwOut{$d$\tcp*[r]{binocular disparity}}
  $I_{n\_{\rm rec}}$      \tcp*[r]{rectified images}
  $I_{n\_{\rm grey}}$     \tcp*[r]{gray images}
  $\{P_n\}$     \tcp*[r]{detected markers' position on $I_{n\_{\rm grey}}$}
  $I_{n\_{\rm markers}}$  \tcp*[r]{images of markers redrawed at the positions}
  $M_{n}$  \tcp*[r]{meshes built by linking markers' centres} 
  
  \While{${\rm i>0}$}{$i=m, \{S_{n\_{0}}\}=\{0\}$  \tcp*[r]{quantity of marker's layers and squence of markers}
  $M_{n\_{\rm i-1}}$, $R_{n\_i}$  \tcp*[r]{updated meshes and extracted ring of edge markers}
  $i=i-1$ 
  $\{S_{n\_{\rm m-i}}\}=\{S_{n\_{\rm m-i-1}}, S_{n\_{\rm m-i}}\}$  \tcp*[r]{updated sequence of markers}
  }
  $d=\{S_{2\_m}\}_x$-$\{S_{1\_m}\}_x$  \tcp*[r]{binocular disparity}
 \end{algorithm}
 
\textbf{Step 1 (Meshing)}: 
Each marker connects to neighboring points with centers, forming two meshes in the left and right images, respectively (\figref{fig_matching}(b)). 
Each internal marker (markers located internally) should have $l=6\times2=12$ links, since connections are reciprocal, whereas the edge markers (markers along the outer edges) should have fewer than 12 links. 

\textbf{Step 2 (Coding)}: 
All edge markers should collectively form an outer edge ring, repeated on the left and right images. These are each sequentially labelled in a circular order, respectively. Each marker in the left image shares the same label with its matching marker in the right image.

\textbf{Step 3 (Iterating)}: 
The edge markers are removed and the internal markers are iterated using the same process as Steps 1 and 2 (\figref{fig_matching}(c)). Throughout the iterative process, the edge ring is continuously updated, and new edge markers are assigned codes until all pairs of markers are completed.

Benefits of this approach include that the arrangement of edge markers in a ring ensures that even markers sharing the same horizontal coordinates will have distinct vertical coordinates, providing each with a unique label. Also, this technique is robust against variations in deformation scale and speed, enabling stable, real-time matching and tracking simultaneously even under large or rapid changes in contact.

\subsubsection{3D Marker Position}

\begin{figure}[!t]
 \vspace{-20pt}
 \centering
 \includegraphics[width=3.3in]{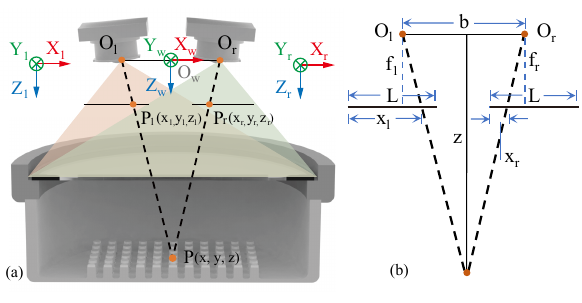}
\vspace{-10pt}
 \caption{Configuration of StereoTacTip. (a) Overview. 
 The world coordinate system $X_w Y_w Z_w$ is located in the middle of the baseline, $X_l Y_l Z_l$ and $X_r Y_r Z_r$ are the camera coordinate systems. 
 (b) Geometric analysis of light rays in StereoTacTip.}
 \label{fig_cord}
 \vspace{0.5em}

 \centering
 \includegraphics[width=3.3in]{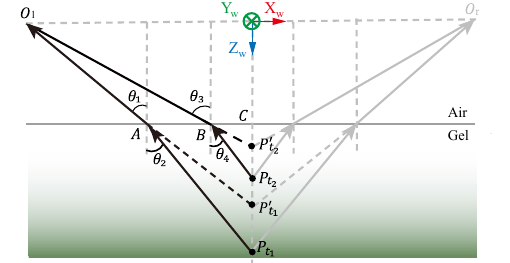}
 \vspace{-5pt}
 \caption{Refraction model for depth correction. 
 $P_{t1}$ is the marker's real initial position, $P'_{t1}$ is the marker's initial position that calculated by the stereo system, $P_{t2}$ is the moved marker’s real position, $P'_{t2}$ is the moved marker position that calculated by the stereo system.  
 $\theta_1$ and $\theta_3$ are the incidence angles, $\theta_2$ and $\theta_4$ are the refraction angles.}
 \label{fig_ray}
 \vspace{-0.4em}
\end{figure}

The disparity is obtained by calculating the difference in the abscissa of each matched point pair, which allows computation of the 3D positions of all markers from the camera's view (as shown in \figref{fig_cord}).

For the calculations, we specify the left and right camera coordinate frames and the world frame: $O_{\rm l}=(X_{\rm l},Y_{\rm l},Z_{\rm l})$, $O_{\rm r}=(X_{\rm r},Y_{\rm r},Z_{\rm r})$ and $O_{\rm w}=(X_{\rm w},Y_{\rm w},Z_{\rm w})$, respectively, where $(x,y,z)\in(X,Y,Z)$ represent coordinate values along these axes. The projections of a marker $P$ in the world frame onto the image coordinate systems are denoted $P_{\rm l}\in O_{\rm l}$ and $P_{\rm r}\in O_{\rm r}$.
The baseline $b$ of the stereo system is the distance between the coordinate origins of the cameras, and is determined through camera calibration. Considering the marker $P(x,y,z)\in O_{\rm w}$, we use the similarity of triangles between $\triangle O_{\rm l} O_{\rm r} P$ and $\triangle P_{\rm l} P_{\rm r} P$ to derive
\begin{align}
 \label{eq_1}
 \frac{b - (x_{\rm r} - x_{\rm l})}{z - f_{\rm l}} &= \frac{b}{z},
\end{align}
where $f_{\rm l}$ is the camera's focal length, 
Therefore,
\begin{align}
\label{eq:2}
z = \frac{b f_{\rm l}}{d},
\end{align}
where $d = x_{\rm r} - x_{\rm l}$ is the binocular disparity. 
Similarly, based on triangle similarity, it can be inferred
\begin{align}
x = \frac{b(x_{\rm l} - x_{\rm c})}{d}, \ \ \ 
y = \frac{b(y_{\rm l} - y_{\rm c})}{d},
\end{align}
where $(x_{\rm c}, y_{\rm c})$ is the position where the camera's principal axis intersects the imaging plane, which can be obtained from calibration. Thus, we can determine the 3D position of the markers by combining the positions from two camera views.

\vspace{-12pt}
\subsection{Surface Reconstruction}
\label{Surface_Reconstruction}

Due to the inherent differences in refractive indices among materials such as acrylic, gel and air, the marker positions calculated in Sec.~\ref{Marker_Localization} will not correspond to the accurate deformation, but rather their virtual positions from the camera’s view. In practise, the light traveling from the marker moves from a medium with a higher refractive index to one with a lower refractive index, causing the actual position of the marker to be further away from that detected by the camera. 

Furthermore, even if we obtain a relatively accurate marker position, it does not represent the position of the skin surface due to the presence of pins connecting the markers to the sensing surface, which have a levering effect that distorts and amplifies the skin deformation. In this section, we address both of these issues by developing a new refraction correction model and applying a marker position correction due to the coupling between the markers and the skin surface.

\vspace{-1.8pt}
\subsubsection{Depth Correction Model}
For simplicity, we consider the acrylic plate and gel together as a single transparent body (\figref{fig_ray}). At the interface between the air and the transparent body, the light path undergoes refraction. Suppose a marker moves from position $P_{t_1}$ at time $t_1$ to position $P_{t_2}$ at time $t_2$. Due to refraction, the camera observes the marker moving from $P'_{t1}$ to $P'_{t2}$. In \figref{fig_ray}, we depict: the incident light rays $OA$ and $OB$, and refracted rays $AP'_{t_1}$ and $BP'_{t_2}$; the angles of incidence $\theta_1$ and $\theta_3$, and the angles of refraction $\theta_2$ and $\theta_4$; and point $C$ is the intersection of the $Z_{\rm w}$-axis of the world coordinate system with the boundary line of the medium.
\begin{figure}[!t]
 \vspace{-18pt}
 \centering
 \includegraphics[width=3.2in]{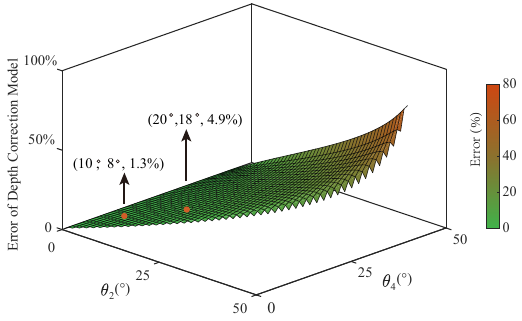}
 \vspace{-6pt}
 \caption{Error analysis of depth correction model. Here we take $\frac{|BC|}{|AC|} = 1.1$, a typical value.}
 \label{fig_e}
 \vspace{-0em}
 \raggedleft
 \includegraphics[width=3.4in]{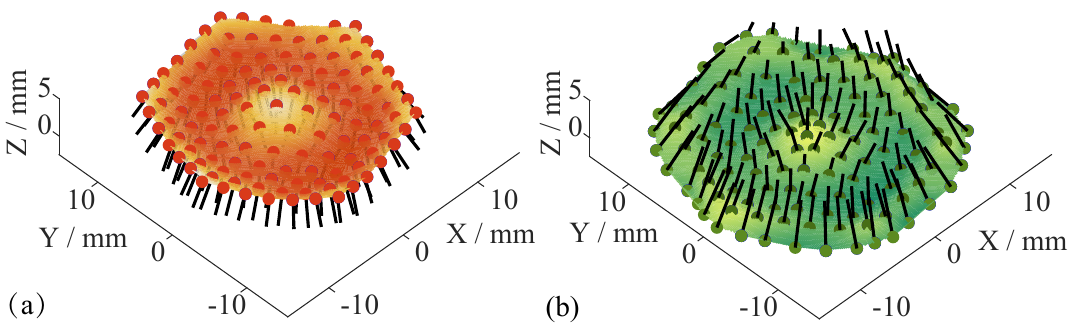}
 \vspace{-3pt}
 \caption{Geometry reconstruction process from marker surface to skin surface. (a) Marker surface and normals. (b) Skin surface and normals.}
 \label{fig_geo}
 \vspace{-0.2em}
\end{figure}

\begin{figure*}[t]
 \vspace{-23pt}
 \centering
 \includegraphics[width=6.5in]{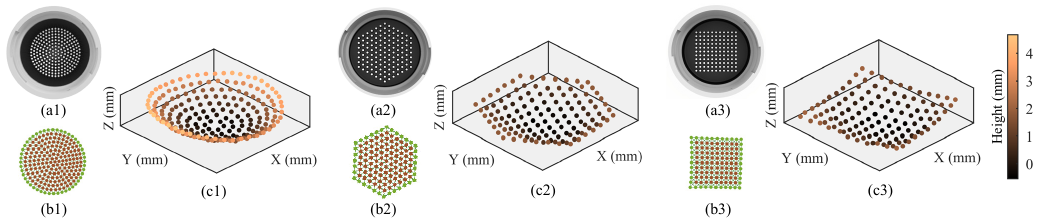}
\vspace{-6pt}
 \caption{Validation of marker localization with different patterns (left: circular, middle: hexagon, right: square). (a1)-(a3) CAD of skin modules. (b1)-(b3) The left meshing of all markers in (a). (c1)-(c3) 3d localization results. Note that the large height variation in (c1) is due to the dome-shaped configuration of (a1).}
 \label{fig_pattern}
 \vspace{-1.5em}
\end{figure*}
Using absolute positions as a ground truth for calibration is challenging because the absolute position of marker points is influenced by factors such as temperature, the amount of glue injected and 3D-printing accuracy. However, since the effect of refraction on object position primarily impacts the $Z$-axis direction~\cite{GelStereoPalm}, we can focus on relative positions instead. By establishing the relationship between actual and virtual position variations---specifically, the relationship between \( |P'_{t_1}P'_{t_2}| \) and \( |P_{t_1}P_{t_2}| \)---we can correct the positions calculated in Eq.~(\ref{eq:2}) of Sec.~\ref{Marker_Localization}. Here, \( |P_{t_1}P_{t_2}| \) can be directly measured, while \( |P'_{t_1}P'_{t_2}| \) is determined through calculation. From the trigonometric relationships shown in \figref{fig_ray}, we derive:
\begin{equation}
\label{eq_3}
\frac{|P_{t_1}P_{t_2}|}{|P'_{t_1}P'_{t_2}|} = \frac{\frac{|AC|}{\tan\theta_2} - \frac{|BC|}{\tan\theta_4}}{\frac{|AC|}{\tan\theta_1} - \frac{|BC|}{\tan\theta_3}}. 
\end{equation}

Then, according to Snell's Law,
\begin{align}
\label{eq_4}
\frac{\sin\theta_1}{\sin\theta_2} = \frac{\sin\theta_3}{\sin\theta_4} = \frac{n_{\text{gel}}}{n_{\text{air}}},
\end{align}
where $n_{\text{gel}}$ and $n_{\text{air}}$ are the refractive indexes of silicone gel and air, respectively. Substituting Eq.~(\ref{eq_3}) into Eq.~(\ref{eq_4}) gives

\begin{align}
\label{eq_5}
\frac{|P_{t1}P_{t2}|}{|P'_{t1}P'_{t2}|} =  \frac{n_{\text{gel}}}{n_{\text{air}}} \cdot \frac{|AC| \sin\theta_4 \cos\theta_2 - |BC| \sin\theta_2 \cos\theta_4}{|AC| \sin\theta_4 \cos\theta_1 - |BC| \sin\theta_2 \cos\theta_3}.
\end{align}

Then, by applying a Taylor expansion to the trigonometric functions in Eq.~(\ref{eq_5}) and retaining terms up to the second order, we obtain after some algebraic simplification
\begin{align}
\frac{|P_{t_1}P_{t_2}|}{|P'_{t_1}P'_{t_2}|} 
&= \frac{n_{\text{gel}}}{n_{\text{air}}} \cdot \left(\frac{|AC| \cdot \theta_4 - |BC| \cdot \theta_2 + O(\theta^3)}{|AC| \cdot \theta_4 - |BC| \cdot \theta_2} \right) \nonumber \\
&= \frac{n_{\text{gel}}}{n_{\text{air}}} \cdot \left(\frac{|AC| \cdot \theta_4 - |BC| \cdot \theta_2}{|AC| \cdot \theta_4 - |BC| \cdot \theta_2} + O(\theta^2)\right) \nonumber \\
&= \frac{n_{\text{gel}}}{n_{\text{air}}} \cdot (1 + O(\theta^2)), \label{eq:final}
\end{align}
where \(O(\theta^2)\) and \(O(\theta^3)\) represent second-order and third-order error terms, respectively. Here Eq.~(\ref{eq:final}) clearly shows that the ratio \(\frac{|P_{t_1}P_{t_2}|}{|P'_{t_1}P'_{t_2}|}\) exhibits a linear relation to the constant ratio of refractive indices. As \(\theta_2\) and \(\theta_4\) become sufficiently small, this asymptotically approaches that constant ratio.

To examine further the error term in Eq.~(\ref{eq:final}), we consider the right-hand side of Eq.~(\ref{eq_5}) and designate it as $\frac{n_{\text{gel}}}{n_{\text{air}}}(1+E)$, then use the MATLAB computer algebra package to find:
\begin{align}
E = \frac{\frac{|AC|}{|BC|} \sin\theta_4 \cos\theta_2 - \sin\theta_2 \cos\theta_4}{\frac{|AC|}{|BC|} \sin\theta_4 \cos\theta_1 - \sin\theta_2 \cos\theta_3}-1.
\end{align}
Results of this error are shown in \figref{fig_e} for a realistic example of $\frac{|BC|}{|AC|} = 1.1$. For example, at values \((\theta_2,\theta_4) = (10^\circ,8^\circ) \), the error is only 1.3\%. The values of \( \theta_2 \) and \( \theta_4 \) are influenced by the design of the stereo system; in this work, the incidence angles do not exceed \(10^\circ\), therefore, the ratio approximates the refractive index ratio with a very small error. Since \( n_{\text{air}} \) is almost 1 under atmospheric pressure (\(n_{\text{air}} = 1.00027\)~\cite{airindex}), this ratio is approximated by \( n_{\text{gel}} \). Thus, given the initial refracted depth $z'$, the corrected depth $z_c$ is
\begin{align}
z_c = z' + |P_{t_1}P_{t_2}| \approx z' + n_{\text{gel}} \cdot |P'_{t_1}P'_{t_2}| = z' + n_{\text{gel}} \cdot z.
\end{align} 


\subsubsection{Skin Surface Reconstruction}

Building on the previous step, we will now correct the 3D position of the skin surface relative to the marker's location. The deformation is assumed to occur under the following conditions:
\begin{enumerate}
    \vspace{-2pt}
    \item The pin remains perpendicular to the skin surface during deformation.
    \item The squeezed gel will not affect the pin's deformation caused by contact.
    \vspace{-2pt}
\end{enumerate}

\uhl{The availability of this assumption relies on the significant stiffness difference between the pin and the gel. This assumption was also utilized in our previous work~\cite{Nicholas2022}. Since the fabrication process and materials remain unchanged, we continue to adopt this assumption.} 

First, we fit a smooth surface \( F_{\rm m}(x, y, z) = 0 \) through the positions of all markers. On \( F_{\rm m} \), we draw the unit normals \(\vec{N}\) from each marker's centre (illustrated in \figref{fig_geo}(a)),
\begin{equation}
\vec{N} = \frac{\vec\nabla F_{\rm m}}{\|\vec{\nabla} F_{\rm m}\|},\ \ \ 
\vec{\nabla} F_{\rm m}=\left( \frac{\partial F_{\rm m}}{\partial x}, \frac{\partial F_{\rm m}}{\partial y}, \frac{\partial F_{\rm m}}{\partial z} \right),
\end{equation}
where \(\|\cdot\|\) is the Euclidean norm. Subsequently, we shift all marker positions in the direction opposite to the normals by a distance $(H+T)$, where $H$ is the height of the pin and $T$ is the thickness of the skin surface. This adjustment results in a new set of points on the skin that correspond to each marker
\begin{equation}
{\vec P}_{\rm s} = (x_{\rm s}, y_{\rm s}, z_{\rm s}) = (x, y, z) - (H + T) \cdot \vec{N},
\end{equation}
then fit a new skin surface \( F_{\rm s}(x_{\rm s}, y_{\rm s}, z_{\rm s}) = 0 \) through all points (illustrated in  \figref{fig_geo}(b)). Therefore, the skin surface is corrected from the marker surface. When the skin is in full contact with the object, \( F_{\rm s} \) can be considered as the contour of the object contact~\cite{2d-3d}. 

\uhl{The sensor's latency from a raw image to a reconstructed image is 283 ms on average.} The reconstruction error of the skin surface will be evaluated in Sec.~\ref{Evaluations}.

\section{Experiments I: Validations}
\label{Validations}
\balance
\subsection{Marker Localization with Different Marker Patterns}
\label{Marker_Localization_with_Different_Patterns}

To verify the adaptability of our DTRC marker localization method  (Sec.~\ref{Marker_Localization}) to various marker arrays, we designed several distinct skin modules: a domed circular pattern of 217   markers, a flat hexagonal pattern of 217 markers, and a flat square pattern of 121 markers, respectively (shown in \figref{fig_pattern}(a1-a3)). For each skin module, the marker localization method was used to reconstruct the 3D positions of the markers. Note that in our DTRC algorithm (Alg. 1), the initial condition \(m\) in step 7 and the number of links \(l\) in the step 8 differ \uhl{according to the values in Table III.}

From the constructed meshes (Fig.~8(b1-b3)), each pattern successfully generates meshes and facilitates marker matching. These successful constructions were achieved despite differences in pattern shapes, variations in the number of edge markers per layer, and changes in marker distances (as observed in the circular pattern). In consequence, the 3D positions of all markers can be obtained (\figref{fig_pattern}(c1-c3)). 
\begin{table}[t!]
\vspace{-2pt}
\centering
\caption{Parameters for marker localization algorithm} 
\begin{tabular}{cccc}
\toprule
Pattern & Circular  & Hexagon  & Square \\
\midrule
Link ($l$) & 12 & 12 & 16 \\
Layer ($m$) & 9 & 7  & 6\\
\bottomrule
\end{tabular}
\vspace{-10pt}
\end{table}

Now that we have established the marker localization method for various marker patterns, we simplify the following treatment by choosing a hexagonal pattern for our remaining experiments. This skin module benefits from a good marker coverage and a regular array; however, we do not expect significant differences with the other modules.

\begin{figure*}[t!]
 \vspace{-18pt}
 \centering
 \includegraphics[width=6.5in]{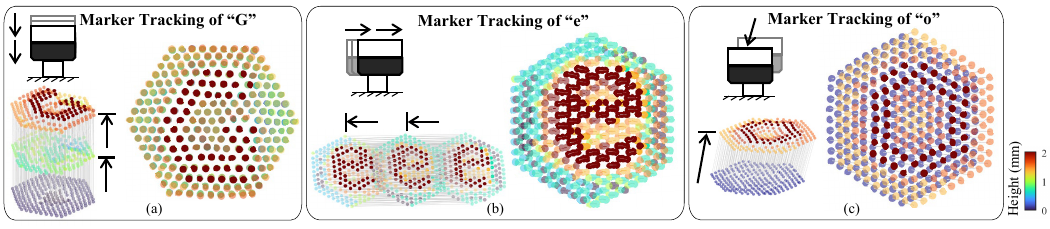}
 \vspace{-5pt}
 \caption{Validation of marker tracking with letters \uhl{(a) ``G'' with vertical motion, (b) ``e'' with horizontal motion and (c) ``o'' with rapid diagonal motion}. 
 In each box, the top left shows the motion diagram, the bottom left displays the 3D visualization of the tracking results, and the right side presents the x-y plane view of the tracking results.}
 \label{fig_tracking}
 \vspace{-2.2em}
\end{figure*}
\begin{figure}[t!]
 \centering
 \includegraphics[width=3.0in]{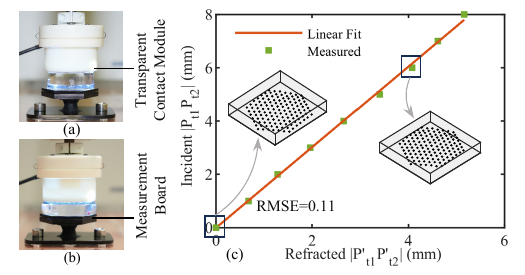}
 \vspace{-8pt}
 \caption{Measurement of refractive index. (a) The moment when touching depth is $0$ mm. (b) The moment when touching depth is $4$ mm. (c) Linear fitting between $|P_{t_1}P_{t_2}|$ and $|P'_{t_1}P'_{t_2}|$.}
 \label{fig_refra}
 \vspace{-2pt}
\end{figure}

\vspace{-10pt}
\subsection{Marker Tracking under Different Motions}
\label{Marker_Tracking}

This section focuses on demonstrating the effectiveness of our DTRC algorithm for marker tracking under various motions.
\hl{We printed the letters ``G'', ``e'' and ``o'' in ABS material and conducted the following three sub-experiments:

\noindent\textbf{Experiment 1 -- Slow vertical motion}: Beginning in contact with the letter ``G'' at contact depth of 1\,mm, we slowly pressed the StereoTacTip downward twice by 1\,mm, capturing three images during the vertical movement.

\noindent\textbf{Experiment 2 -- Horizontal motion}: Starting in contact with the letter ``e'' at a contact depth of 3\,mm, we translated the StereoTacTip laterally right twice by 1\,mm, capturing three images during the horizontal movement.

\noindent\textbf{Experiment 3 -- Rapid diagonal motion}: Starting just in contact with the letter ``o'' at contact depth $\sim$0\,mm, we rapidly moved the sensor down and left by 2\,mm in each direction, capturing images at the start and end positions.

Overall, marker tracking was highly effective across the various types of movement (\figref{fig_tracking}). In scenario (a), where the markers had small displacements between adjacent frames, traditional methods such as finding the nearest points in the $xy$-plane (\textit{e.g.}, KNN) can also effectively track the markers. For scenario (b) with large marker displacements and scenario (c) with very large rapid displacements, distinguishing adjacent points in the $xy$-plane became challenging. The advantage of our DTRC algorithm then becomes evident: despite large rapid displacements that could cause marker confusion, marker tracking was accurately maintained.}


\vspace{-5pt}
\subsection{Coefficient of Refraction Measurement for Calibration}
\label{Refractive_Model}

An issue for the determination of marker and skin displacement is that the refractive index of the transparent gel layer cannot be directly measured. To overcome this, we calculate $n_{\rm gel}$ by measuring directly the positions $|P_{t_1}P_{t_2}|$ and $|P'_{t_1}P'_{t_2}|$ in Eq.~(\ref{eq:final}) to use for calibration.

For calibration, we printed a measurement board with the same marker arrangement as the 127-marker flat hexagon skin module, using Verowhite and Veroblack (the same material as Verowhite except for color), and secured it to the robotic operation platform (\figref{fig_refra}(a,b)). We then used molds to create a transparent skin module without skin, pins and markers and mounted it as the end effector of the desktop robot arm.

The calibration process was initialized by moving the end effector until the gel just fits the measuring board (position as shown in \figref{fig_refra}(a)), which was set as the initial position $z_0 = 0$\,mm. 
Using our marker localization method (Sec.~\ref{Marker_Localization}), we calculated the 3D positions of all the markers and their average depth, $\bar{z}_1$. Next, we incrementally moved the end effector in 1\,mm steps up to 8\,mm (to avoid damage), calculating $\bar{z}_n$ at each step (example at 4\,mm shown in \figref{fig_refra}(b)). The distance moved by robot arm, $|P_{t1}P_{t2}|$, served as a ground truth for distance changes, while the distance variation, $|P'_{t1}P'_{t2}|$,  was calculated from the camera’s view as $\Delta z = \bar{z}_n - z_0$. The results were then averaged over 5 repeated trials.

From a plot of the measured values (\figref{fig_refra}(c)), the ratio between $|P_{t1}P_{t2}|$ and $|P'_{t1}P'_{t2}|$ is highly linear, with a fit $n_{\rm gel}=1.51$. According to the literature, the refractive indices of StereoTacTip's refractive layer are 1.57~\cite{157} for the acrylic board 1.40~\cite{140} for the elastomer. Our test results closely align with these actual values, falling reasonably between them. 

\vspace{-2pt}
\section{Experiments II: Evaluations}
\label{Evaluations}
\balance

For evaluation, we separated the experiments into two groups according to internal factors (\textit{i.e.}, the design parameters of StereoTacTip) and external factors (\textit{i.e.}, the parameters of the contacted objects), to explore their impact on the reconstruction performance. 

\begin{figure*}[!t]
 \vspace{-18pt}
 \centering
 \includegraphics[width=6.6in]{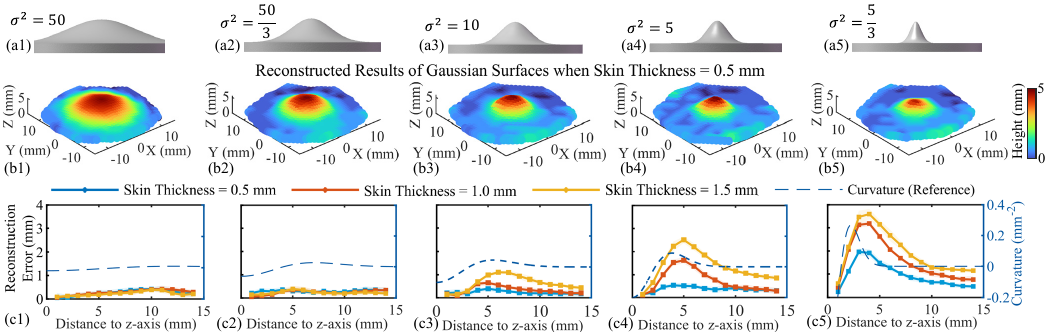}
 \vspace{-3pt}
 \caption{Reconstruction results of Gaussian surfaces with different skin thickness to test the dependence of skin thickness and object sharpness. First row: CAD of Gaussian surfaces. Second row: reconstructed results of Gaussian surfaces when skin thickness = 0.5 mm. Third row: comparison of reconstruction Error when skin thickness varies. The curvature of Gaussian surfaces serves as a reference. Every column is with the same $\sigma^2$.}
 \label{fig_gaussian}
\vspace{-1.55em}
\end{figure*}
%
\vspace{-12pt}
\subsection{Dependence of Skin Thickness and Object Sharpness}
\label{Performance_of_Skin_Thickness_and_Object_Sharpness}

In contact-based reconstruction, a critical factor influencing accuracy is the degree of conformity between the sensor and the object. 
To explore this issue, we utilized three skin modules with varying skin thicknesses (0.5\,mm, 1.0\,mm, 1.5\,mm) to reconstruct differently curved surfaces.

We consider the reconstruction of Gaussian surfaces \(z = h\,\exp({-(x^2 + y^2)/2\sigma^2})\) of equal height $h=5$\,mm and distinct widths \( \sigma^2 = \{50, 50/3, 10, 5, 50/3\}\)\,mm$^2$ (shown in \figref{fig_gaussian}(a1-a5)). In all cases, StereoTacTip was pressed downward by 5\,mm from its initial position at a speed of 1\,mm/s to contact the surface, with this process repeated three times, rotating the sensor 120 degrees between each repetition. 

Let us first consider the reconstruction results for a skin thickness of 1.0\,mm (\figref{fig_gaussian}(b1-b5)). Evidently, when \( \sigma^2 \geq 10 \), the reconstructed surface closely approximates the original model surface. However, at \( \sigma^2 = 5 \), the reconstructed surface begins to deviate from the original surface in regions with significant curvature changes. At \( \sigma^2 = 5/3 \), the reconstructed surface fails to capture the sharpness of the original surface, rendering the reconstruction ineffective in regions with high curvature. To quantify this observation, we calculated the reconstruction error at distances from Gaussian centre (plotted in \figref{fig_gaussian}(c1-c5)). 


Overall, the distribution trend of reconstruction errors closely resembles the curvature of the corresponding Gaussian surface (reference curvature shown in \figref{fig_gaussian}(c1-c5) as dashed blue curve). We interpret this as due to a lower ability to conform to surfaces of the soft skin as it increases in thickness. As the curvature difference increases, the surfaces fail to fit closely, resulting in gaps that lead to increased reconstruction errors. Additionally, the transition from non-conformity to conformity requires gradual adaptation. Therefore, the extreme values in (c4) and (c5) have a significant spatial lag. As the skin thickness increases, the errors for the same Gaussian surface also increase, indicating that thinner skins have better deformation adaptability. Therefore, we opted for a skin thickness of 0.5\,mm in subsequent experiments.

\vspace{-13pt}
\subsection{Dependence on Marker Density and Object Spatial Frequency}
\label{Performance_of_Marker_Density_and_Object_Frequency}


When the skin contacts over multiple separated points, rather than just a single smooth shape as in the previous section, additional challenges are introduced into the shape reconstruction. This difficulty arises because the skin surface, when compressed, may not conform to the lowest points of the object shape (\textit{e.g.}, valleys in a wave). Additionally, a low marker density could worsen this issue.

Therefore, we fabricated three skin modules with different marker separations (sparse 3.54\,mm, moderate 2.54\,mm, dense 1.80\,mm) to reconstruct sine wave surfaces $z = 2.5\sin(\omega x)$ of distinct frequencies \(\omega=\{\pi/15,2\pi/15,\pi/5,4\pi/15,2\pi/5\}\), as shown in \figref{fig_sine}(a1-a5). For the experiments, the StereoTacTip was pressed 5\,mm onto the objects at a speed of 1\,mm/s. Each experiment was repeated three times, with StereoTacTip rotated 120 degrees between each repetition. 

Let us first consider the results with markers of moderate density (\figref{fig_sine}((b1-b5)). Evidently, as surface spatial frequency increases, the reconstruction becomes more difficult up to \(\omega=2\pi/5\), where the sensor could no longer accurately distinguish the ridges of the sine surface (\figref{fig_sine}((b5)). We interpret this as due to, at this frequency, the distance between the extremes of the sinusoidal surface being less than twice the marker spacing, which prevents markers from being placed at the deepest points of the gaps. This observation is further confirmed by a cross-sectional plot (\figref{fig_sine}(c)), which also shows the sensor skin could no longer touch the lower half of the sinusoidal surface, rendering the reconstruction of the lower half ineffective. To quantify this observation, we calculated the reconstruction error under different densities and analyzed it in two aspects.

Overall, as the surface spatial frequency increases, the conformity of the sensor skin to the surface (\textit{i.e.}, reconstruction error) gradually deteriorates. Since the surface is preferentially contacted during the process, the consistency of the reconstruction on the upper surface is better than that on the lower. Figure~\ref{fig_sine}(d) separately calculates the reconstruction error of the upper surface. An evident trend is observed where the denser marker arrangement results in smaller errors compared to the sparser arrangement, which is most pronounced close nearby \(\omega = \pi/5\). Otherwise, the advantage of the denser marker arrangement appears to diminish, which we attribute either to the skin conforming well to the surface for \(\omega \ll \pi/5\), or the increasing density of markers being unable to compensate for the sensor skin's inability to touch the surface for \(\omega \gg \pi/5\).

This finding is further substantiated by the reconstruction results of the lower surface. Rather than calculating the overall error for the lower surface, we consider the difference between the lowest point of the reconstructed lower surface and the actual surface (\figref{fig_sine}(e)). This approach more accurately reflects the degree to which the lower surface deviates from the intended reconstruction. When \(\omega \leq \pi/5\), the error remains within 2.5\,mm, which corresponds to half the height of the sinusoidal surface. However, when \(\omega > \pi/5\), the error exceeds 2.5\,mm, suggesting that the skin can no longer conform to any point on the lower surface. This suggests that the tension in the sensor skin hinders its ability to conform to the lower surface.

Since a denser marker arrangement yields better results, we opted for the skin module with dense markers for the experiments in the next section.

\vspace{-0.5em}
\section{Experiments III: Demonstrations}
\label{Demonstrations}
\balance

Tactile sensors are rarely used for surface reconstruction in scenarios that consider complex and varied local and global topography characterized by significant depth changes of several mm or more. We attribute this as due to the challenges of requiring a highly compressible tactile sensor while also being able to reconstruct in 3D. The StereoTacTip can handle this reconstruction task, which we use as a benchmark for evaluating performance, as we describe below.

\vspace{-12pt}
\subsection{Model Generation}

A 3D topographic map of the world was created by QGIS, an open-source geographic information system. To generate the earth topographic global digital elevation model (DEM), firstly, we downloaded a 30 arc-second resolution DEM from ~\cite{ETOPO2022} and imported it into QGIS in raster format. Subsequently, taking the coastline as the demarcation line, we extracted the DEM of global landmass by the geospatial data abstraction library~\cite{GDALlib2024},  and reset the DEM of the ocean floor to a fixed value of $-3000$ meters. Finally, the DEM of global landmass was reprojected to EPSG:4326 coordinate reference system and converted to STL format with a model size of 320$\times$160$\times$9.7\,mm$^3$. The STL file enabled us toprint the map in ABS material to use for our demonstration experiments.

\vspace{-12pt}
\subsection{Experimental Procedures}

First, we secured the map to the operation platform, ensuring the entire model remained within the workspace of the desktop robot arm equipped with a StereoTacTip (dense markers, skin thickness = 0.5\,mm) as end effector. We establish a global coordinate system $O_{\rm {global}}=(X_{\rm {global}},Y_{\rm {global}},Z_{\rm {global}})$ with the origin at the position of (180$^\circ$E, 90$^\circ$S, 2\,mm), corresponding to an altitude of 0\,m in the real-world topographic map. The positions of markers are then transformed to the global coordinates using a transformation matrix $r^{\text{global}}_{\text{marker}} = r^{\text{w}}_{\text{marker}} \cdot r^{\text{global}}_{\text{w}}$.

\begin{figure*}[!t]
 \vspace{-18pt}
 \centering
 \includegraphics[width=6.5in]{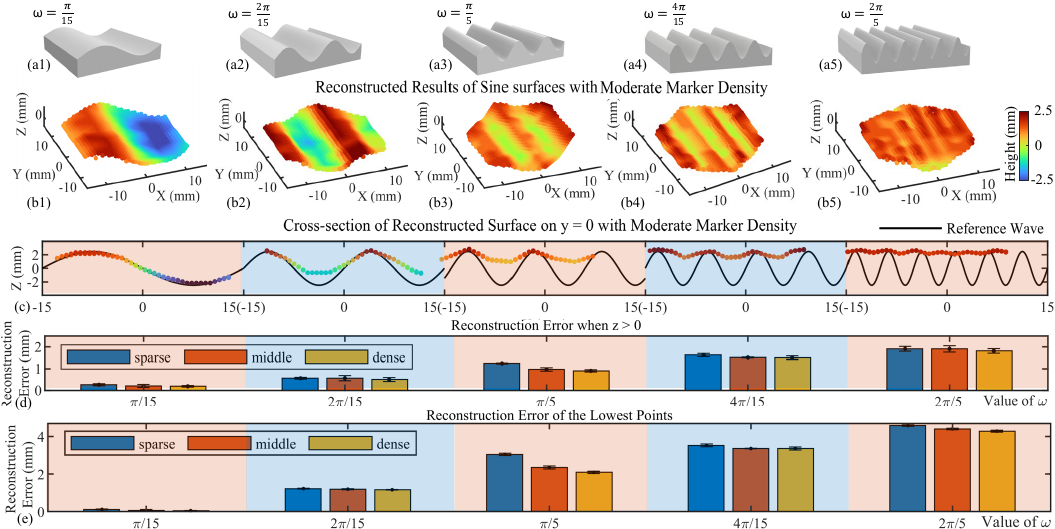}
\vspace{-5pt}
 \caption{Reconstruction results of Sine surfaces with different marker densities to test the dependence on marker density and object spatial frequency. First row: CAD of reconstructed Sine surface. Second row: examples of reconstructed results with middle marker density. Third row: slice view of the results in the second row when y = 0. Fourth row: comparison of reconstruction error when marker density varies in the case where $z > 0$. Last raw: comparison of reconstruction error of the lowest points (such as valleys in a wave) when marker density varies. Every column is with the same $\omega$.}
 \label{fig_sine}
 \vspace{-1.55em}
\end{figure*}

Next, we preset nine representative terrain areas in the model, and calculated the range of their positions in global coordinates for use in controlling the desktop robot arm. For each region, the robot moved its end effector to reconstruct these areas in a zigzag scanning route with 15\,mm horizontal steps (\figref{fig_map}(b)). During each cycle, the StereoTacTip moved vertically downward to tap at a position $(x_{i},y_{i},0)$\,mm from an initial position $(x_{i},y_{i},8)$\,mm, then returns back to its initial height. This height is set to prevent interference between the StereoTacTip and the highest points of the topographic model. After all points in the region have been scanned, we obtain a set of stereo tactile images, which are processed into sets of marker positions ${R_i} =\{(x_{i,m}, y_{i,m}, z_{i,m})\}_{{m}=1}^{n_{\rm m}}$, where $n_{\rm m}$ is the number of markers. These marker positions are combined to give all points spanning the mapped region $R=\{R_i\}_{i=1}^{n}$, where $n$ is the number of contacts.

\vspace{-12pt}
\subsection{Data Processing}

Due to the robot's translational movement in the $(x,y)$-plane being smaller than the diameter of the marker coverage, there is substantial overlap between local patches of the map from each tap. These overlaps occur on the outer part of the contact region, where we observed a bias that ${z_{i,m}}$ is increasingly overestimated towards the periphery (because the skin is less compressible where it joins with the rigid mount). Therefore, we adopt an interpolation method that smooths the points with lower ${z_{i,m}}$ values in the overlap region (\figref{fig_map}(a)), with implementation as follows:

\subsubsection{Identify Overlapping Regions}
First, the overlap region is identified from the density of points in the $(x,y)$-plane, which will be higher on the overlap region. For each point $(x_{i,m},y_{i,m})$, we found its nearest-neighbour distance $d_{i,m}$, then applied a threshold of $T=0.6$\,mm to classify whether it is in the overlap region ($d_{i,m}<T$) or not. The collection of these overlapping points we designate as $R_{\text{overlap}}\subset R_i \cup R_j$ between two overlapping regions $R_i$ and $R_j$, and those above the threshold are denoted as $R_{\text{nonoverlap}}\subset R_i\cup  R_j$.

\subsubsection{Extract Contiguous Region} From $R_{\text{overlap}}$, we remove those point that are biased towards higher ${z_{i,m}}$ on the periphery. This is done by pairing the nearest neighbouring points, and selecting the one with the lowest ${z_{i,m}}$, the collection of which we denote $R_{\text{lower}}\subset R_{\text{overlap}}$.

\subsubsection{Merge Overlapping Regions}
These lower points and those in the non-overlapping region are merged to create a new contiguous set of points $R_{\text{lower}} \cup R_{\text{nonoverlap}}\subset R_i\cup R_j$. This process is repeated over all $n$ regions $R_i$ to build up a new contiguous point map over the entire region $R_{\rm new}\subset \{R_i\}_{i=1}^n$.

\subsubsection{Smooth Point Map}

A spatial smoothing operation is then applied over the $z$-values $Z_{\rm new}$ of the points in $R_{\rm new}$ to reduce noise, for which we use a convolution operation
\begin{equation}
\Phi_\epsilon Z_{\rm new} = Z_{\rm new} * \phi_\epsilon.
\end{equation}
The convolution operator is defined by a mollifier function defined over the unit disk in the $(x,y)$-plane 
\begin{equation}
\label{eq_12} 
\phi(\rho) = 
\begin{cases} 
\frac{1}{I} \exp \left( - \frac{1}{1 - \rho^2} \right),&\ \  0\leq \rho < 1, \\
0, &\ \  \rho \geq 1,
\end{cases}
\end{equation}
where $\rho=\sqrt{x^2 + y^2}$ and the normalization factor $I$ sets $\iint \phi(\rho(x,y))\, {\rm d}x \, {\rm d}y=1$. It is conventional to use a length scale $\epsilon$ in Eq.~(\ref{eq_12}), such that
\begin{equation}
\phi_\epsilon(\rho) = \frac{1}{\epsilon^2} \cdot \phi \left( \frac{\rho}{\epsilon} \right),
\end{equation}
which is then used in the convolution operation. Overall, this approach yields a smoother result for the topographic model reconstruction (here using a value $\epsilon=0.25$\,mm). 

\vspace{-12pt}
\subsection{Results and Analysis}

\begin{figure*}[!t]
 \vspace{-18pt}
 \centering

 \centering
 \includegraphics[width=7.13in]{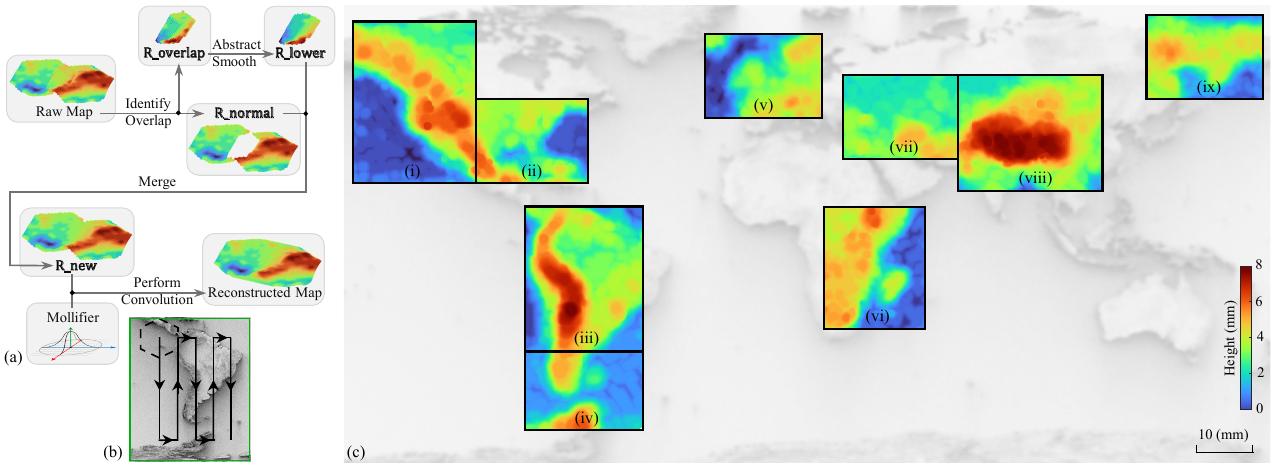}
 \caption{3D reconstructed results of typical topographic areas. The areas represented by the letters in the diagram are (i) the Rocky Mountains, (ii) the Gulf of Mexico, (iii) the Andes Mountains, (iv) Drake Strait, (v) the English Channel, (vi) Mozambique Strait, (vii) the Black and Caspian Seas, (viii) the Himalaya Mountains and (ix) the Sea of Okhotsk.}
 \label{fig_map}
 \vspace{-1.55em}
\end{figure*}

The reconstruction results are shown in \figref{fig_map}(c). Overall, we can clearly see the contours of the continents along with the spatial distribution and differences in the terrain. We now select nine representative regions for further analysis.

{\em 1) Typical mountains: (i) \textbf{Rocky Mountains}; (iii) \textbf{Andes Mountains}; (viii) \textbf{Himalaya Mountains}}: According to~\figref{fig_map}(c-iii), we have reproduced the narrow and elongated characteristics of the Andes Mountains. However, due to the steepness near the highest peaks, the StereoTacTip struggled to accurately reconstruct regions with significant curvature transitions from the base to the summit. This results in a situation similar to the narrow Gaussian in~\figref{fig_gaussian}(b5), making the reconstructed area wider than the original model of the mountain range. On the other hand, the Himalayas have relatively gentle terrain changes, which allows our reconstruction to better maintain the original shape. In comparison, the terrain of the Rocky Mountains is less rugged, resulting in a more accurate reconstruction range.

{\em 2) Typical sea areas: (ii) \textbf{Gulf of Mexico}; (vii) \textbf{Black and Caspian Seas}; (ix) \textbf{Sea of Okhotsk}}:  The Sea of Okhotsk is a large marginal sea which on the scale of the model in \figref{fig_map}(c-ix) has a maximum width of 17\,mm. This is close to the difference between the two extreme values of the sinuosoid reconstruction in~\figref{fig_sine}(c) when $\omega=\pi/15$, and so is able to accurately reconstruct the base of the curve. When the marine area is reduced to the scale of the Gulf of Mexico model, it is still possible to reconstruct the overall contour, but we expect there are inaccuracies in the estimated depths at the bottom of the bay, similarly to the sinuosoids in \figref{fig_sine}(b2-b4). Lastly, the Black and Caspian Seas have a model width of just 2\,mm, leading to an inaccurate reconstruction analogous to the sinuosoids in~\figref{fig_sine}(b5) when $\omega=2\pi/5$.


{\em 3) Typical channels: (iv) \textbf{Drake Strait}; (v) \textbf{English Channel}; (vi) \textbf{Mozambique Strait}:} Connecting South America and Antarctica, the Drake Strait has a width of 9\,mm in the model, similar to the sinuosoid in \figref{fig_sine}(b1) with $\omega = \pi/15$. As shown in~\figref{fig_map}(c-iv), the reconstructed continents at both ends of the strait and the bottom of the strait can be clearly distinguished. The reconstruction of the Mozambique Channel is less clear. Although the gap between the island and the continent can be identified, the depth construction of the strait failed, similar to the sinuosoid in~\figref{fig_sine} where $\omega > 2\pi/15$. Likewise, a similar situation occurred for the English Channel, which was only just discernible due to being a small feature on the topographic map.  


\section{Conclusions and Future Works}
\label{Conclusions}

\uhl{In this paper, we proposed a modular marker-based stereo-tactile sensor called the StereoTacTip designed for 3D geometry reconstruction. We found that the stereo camera setup significantly enhanced the interpretability and reliability of the reconstructed depth profile, employing an innovative marker matching and tracking algorithm coupled with an optical depth correction model. We also introduced an analytic model to reconstruct skin deformation from the 3D marker displacements that applies to complex biomimetic marker-skin arrangements. For validation, this  was then applied to reconstruction of a topographic terrain map. A key advantage of our approach is its robustness, in that it maintains its effectiveness even with variations in pattern shapes, marker numbers, and marker morphologies. Furthermore, many of the contributions should also improve the performance of simpler marker-based VBTSs. Overall, our work illustrates that a thorough understanding and evaluation of the marker-based VBTS principles are crucial for enhancing their performance in 3D reconstruction and beyond.}

\uhl{We conclude by indicating some limitations and directions for future work. First, this paper only explored spatial geometry, and did not consider other tactile features such as force that are important for dexterous manipulation~\cite{ford2025tro}. Our expectation is that the principles developed here will extend more generally to modeling marker-based VBTS outputs, which is an important open topic for investigation particularly for shear forces~\cite{John}. Second, the StereoTacTip developed here was based on a standard 40\,mm dia. TacTip module~\cite{TacTipFamily}, which is larger than more recent fingertip-sized modules for mounting on robotic grippers and hands~\cite{DexiTac,li2025ijrr,ford2025tro}. However, we expect that the continuing miniaturization of camera technology will make it much easier to fit multiple cameras in ever-smaller VBTS form factors~\cite{2d-3d}. This progress will help VBTSs become ubiquitious for robot touch, and make it standard practice to use multiple cameras within tactile sensor modules.}

\bibliographystyle{IEEEtran}
\footnotesize{\bibliography{references/reference.bib}}

\vspace{-3.5em}
\begin{IEEEbiography}
  [{\includegraphics[width=0.8in,clip,  keepaspectratio]{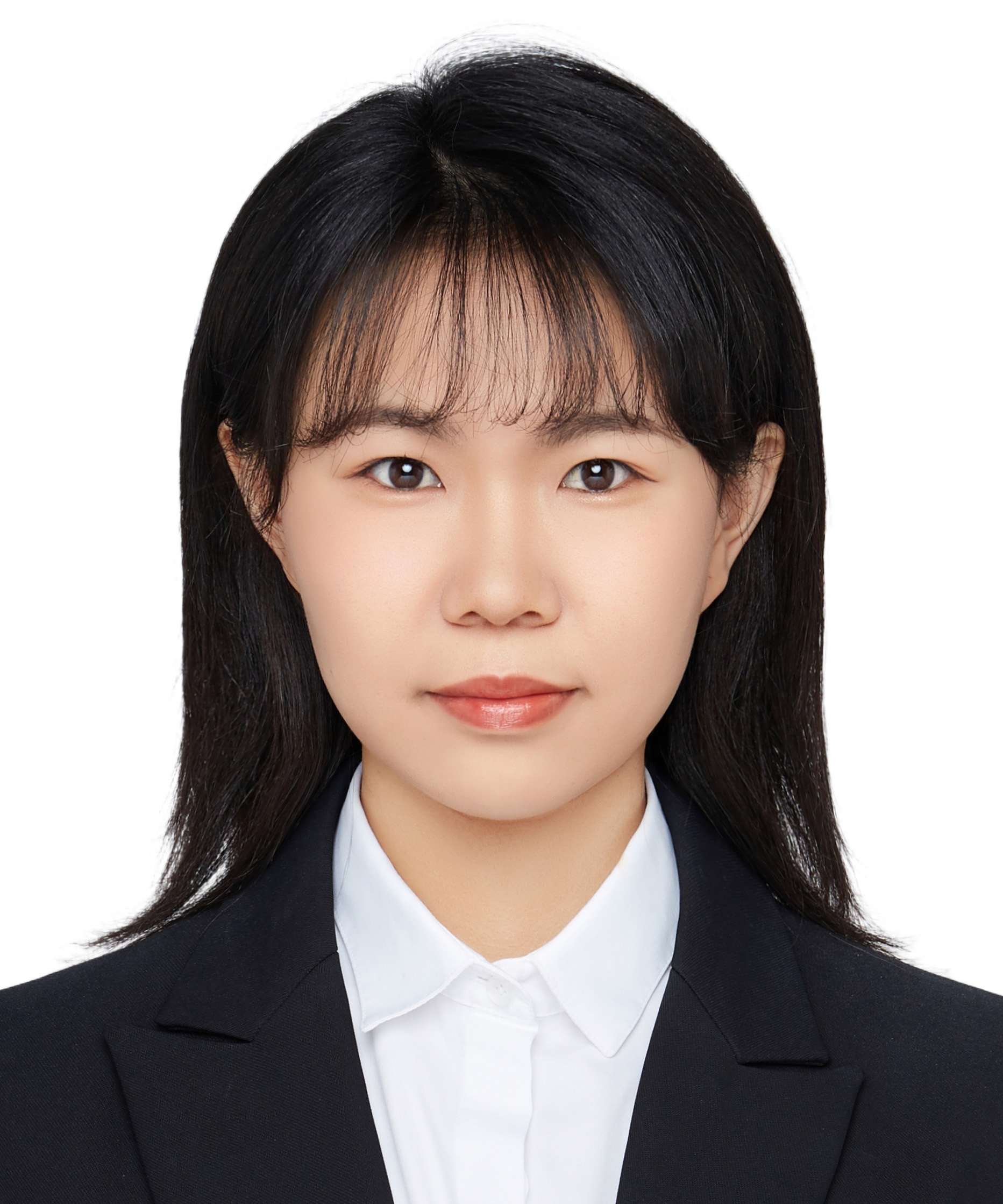}}]
  {Chenghua Lu}
  received the B.S. degree in Mechanical Engineering from Northeastern University, Shenyang, China, in 2017, and the M.S. degree in Mechanical Manufacturing and Automation from the University of Chinese Academy of Sciences, Beijing, China, in 2021. She is currently working toward the Ph.D. degree majoring in Engineering Mathematics with the School of Mathematics Engineering and Technology and Bristol Robotics Laboratory, University of Bristol, Bristol, UK. Her research interests include tactile sensing and soft robotics. 
\end{IEEEbiography}

\vspace{-3em}
\begin{IEEEbiography}
  [{\includegraphics[width=0.8in,clip,  keepaspectratio]{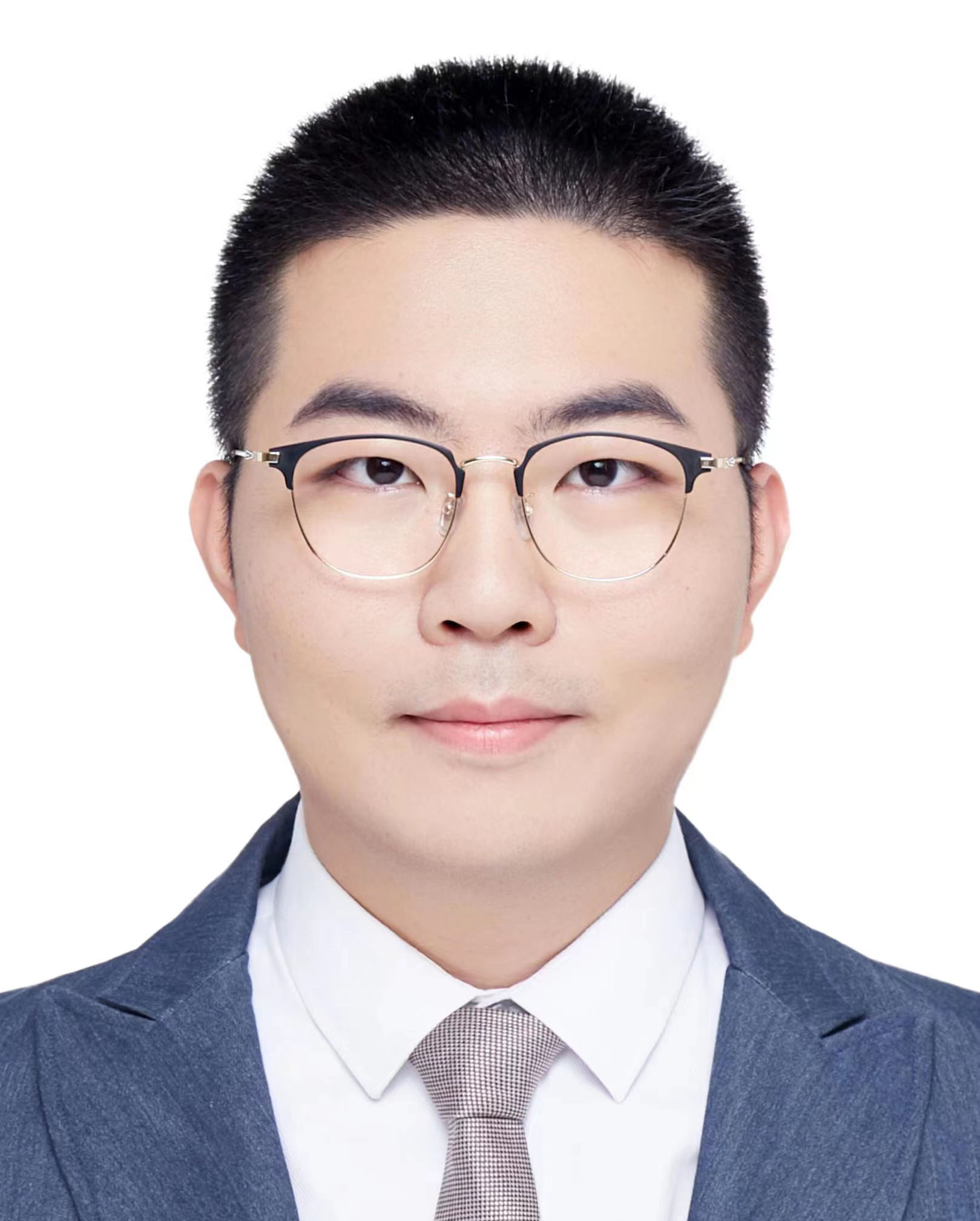}}]
  {Kailuan Tang}
  received a B.S. degree in Communication Engineering from the Southern University of Science and Technology (SUSTech), Shenzhen, China in 2017. He is currently working towards a Ph.D. degree majoring in Mechanics with the School of Mechatronics Engineering, Harbin Institute of Technology. His research interests include underwater robots, biomimetic control, and soft robot perception.
\end{IEEEbiography}

\vspace{-4em}
\begin{IEEEbiography}
  [{\includegraphics[width=0.8in,clip,  keepaspectratio]{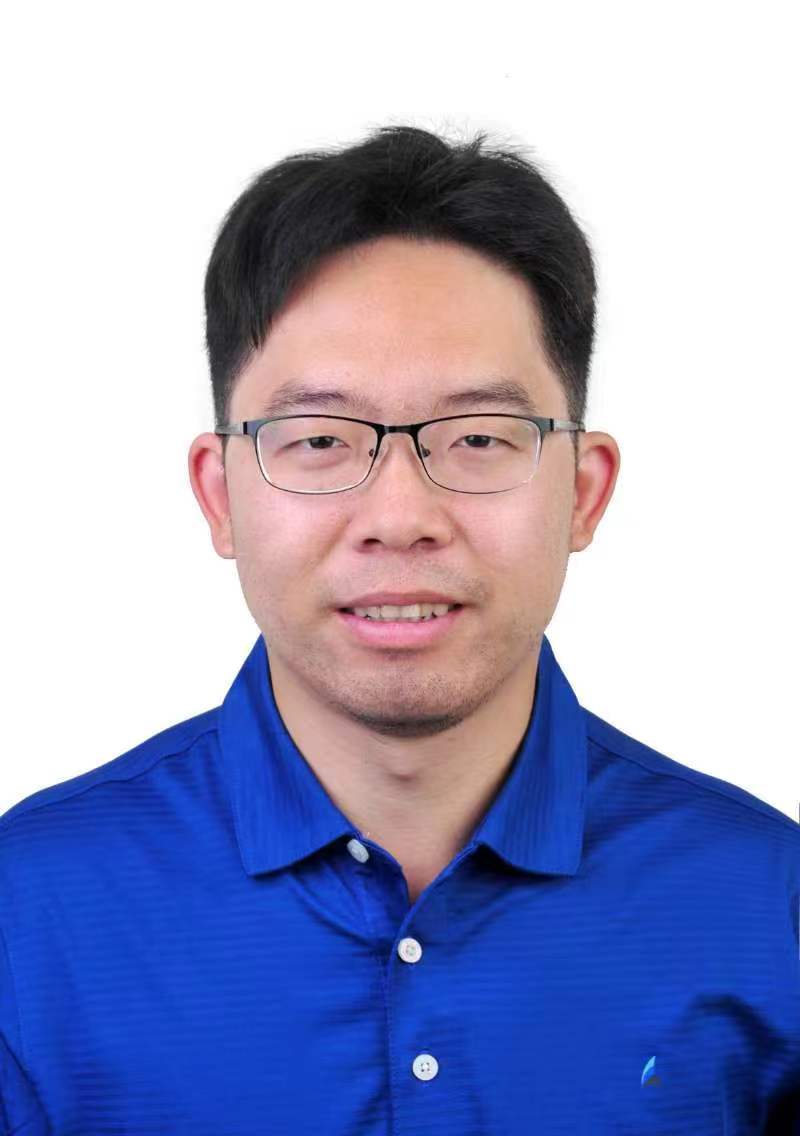}}]
  {Xueming Hui}
  received a B.S. degree in Mathematics and Applied Mathematics from the Southern University of Science and Technology (SUSTech), Shenzhen, China, in 2017, and Ph.D degree in Mathematics from Brigham Young University, Provo, U.S.A., in 2023. He is currently an Assistant Professor at Fuyao Institute for Advanced Study, Fuyao University of Science and Technology, Fuzhou, China. His research interests include dynamical systems and topology. 

\end{IEEEbiography}

  \vspace{-4em}
\begin{IEEEbiography}
  [{\includegraphics[width=0.8in,clip,  keepaspectratio]{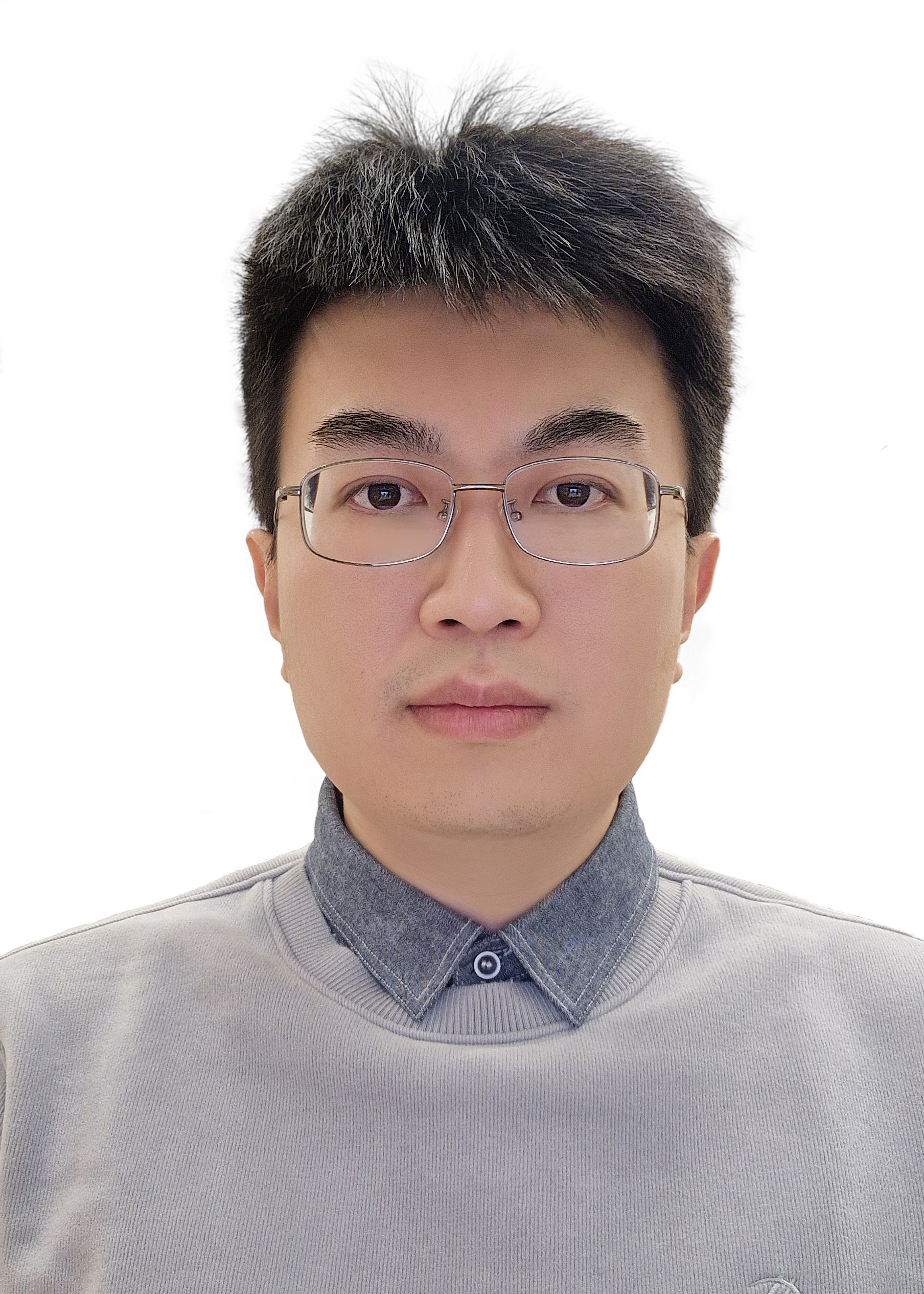}}]
  {Haoran Li}
   received his B.E. in Vehicle Engineering from Wuhan University of Technology (China), followed by an M.Sc. in Robotics (2020) and PhD in Engineering Mathematics (2025) from the University of Bristol, UK. He is currently an Assistant Professor in the School of Robotics at Xi'an Jiaotong-Liverpool University (China). He previously served as a postdoctoral researcher at Bristol Robotics Laboratory, University of Bristol. His research focuses on tactile sensor development and electromechanical systems for dexterous robotic hands.
\end{IEEEbiography}

\vspace{-4em}
\begin{IEEEbiography}
  [{\includegraphics[width=0.8in,clip,  keepaspectratio]{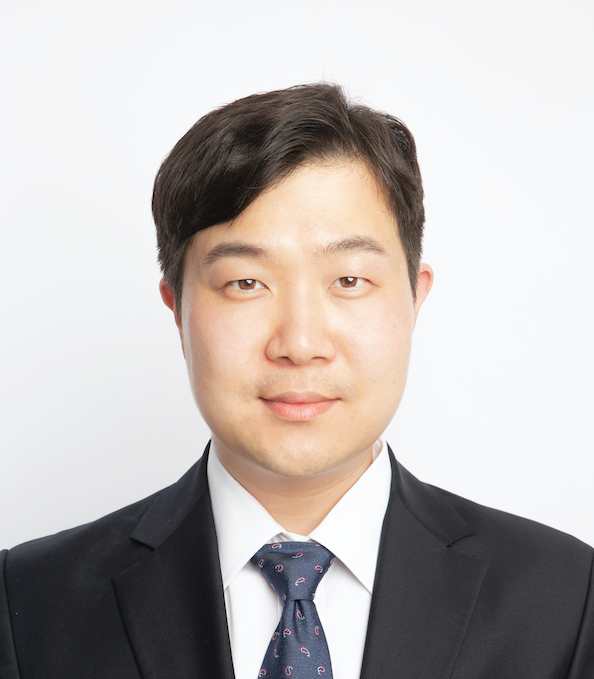}}]
  {Saekwang Nam}
  received a B.E. degree in Human $\&$ Mechanical Engineering from Kanazawa University, Japan, in 2011, and a Dr.rer.nat. degree in Computer Science from the University of T\"ubingen, Germany, in 2022. After working as a postdoctoral researcher at Bristol Robotics Laboratory and the University of Bristol, he is currently an Assistant Professor in the Graduate School of Data Science at Kyungpook National University, Daegu, South Korea. His research interests include tactile sensing and the development of tactile sensors for robots.
\end{IEEEbiography}

\vspace{-3em}
\begin{IEEEbiography}
  [{\includegraphics[width=0.8in,clip,  keepaspectratio]{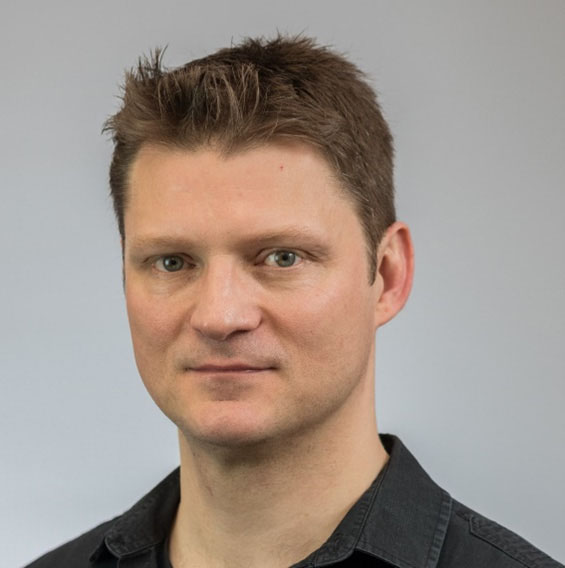}}]
  {Nathan F. Lepora} received the B.A. degree in Mathematics and the Ph.D. degree in Theoretical Physics from the University of Cambridge, Cambridge, U.K. He is currently a Professor of Robotics and AI with the University of Bristol, Bristol, U.K. He leads the Dexterous Robotics Group in Bristol Robotics Laboratory with research funding from Horizon Europe, Leverhulme Trust, EPSRC and The Royal Society. This group won the ‘University Research Project of the Year’ at the 2022 Elektra Awards.
\end{IEEEbiography}

\end{document}